\documentclass[10pt,twoside]{article}

\usepackage{times}
\usepackage[utf8]{inputenc}
\usepackage[T1]{fontenc}
\usepackage{graphicx}
\usepackage[fleqn]{amsmath}
\usepackage{footnote}

\usepackage{taln2017}
\usepackage[frenchb]{babel}
\usepackage{tabularx}
\usepackage{multirow}
\usepackage{hyperref}

\title{Predicting the Semantic Textual Similarity\\with Siamese CNN and LSTM}

\author{Elvys Linhares Pontes\up{1}\quad Stéphane Huet\up{1} \quad Andréa Carneiro Linhares\up{2} \quad Juan-Manuel Torres-Moreno\up{1,3} \\
  {\small
    (1) LIA, Université d'Avignon et des Pays de Vaucluse, Avignon, 84000 France \\ 
    (2) Universidade Federal do Ceará, Sobral, Ceará Brazil \\ 
    (3) \'Ecole Polytechnique de Montréal, Montréal, Canada \\
    \texttt{
      \{elvys.linhares-pontes, juan-manuel.torres, stephane.huet\}@univ-avignon.fr\\
         andrea.linhares@ufc.br 
}}}

\begin{document}
\maketitle

\resume{}{
La Similarité Textuelle Sémantique (STS) est la base de nombreuses applications dans le Traitement Automatique du Langage Naturel (TALN).
Notre système combine des réseaux neuronaux convolutifs et récurrents pour mesurer la similarité sémantique des phrases.
Il utilise un réseau convolutif pour tenir compte du contexte local des mots et un LSTM pour prendre en considération le contexte global d'une phrase.
Cette combinaison des réseaux préserve mieux les informations significatives des phrases et améliore le calcul de la similarité entre les phrases.
Notre modèle a obtenu de bons résultats et est compétitif avec les meilleurs systèmes de l'état de l'art.
}

\abstract{}{
Semantic Textual Similarity (STS) is the basis of many applications in Natural Language Processing (NLP).
Our system combines convolution and recurrent neural networks to measure the semantic similarity of sentences.
It uses a convolution network to take account of the local context of words and an LSTM to consider the global context of sentences.
This combination of networks helps to preserve the relevant information of sentences and improves the calculation of the similarity between sentences.
Our model has achieved good results and is competitive with the best state-of-the-art systems.
}

\motsClefs
  {Similarité, Réseaux de neurones siamois, LSTM, CNN}
  {Similarity, Siamese Neural Networks, LSTM, CNN}

\begin{otherlanguage*}{english}

\vspace{-0.5cm}
\section{Introduction}

Semantic Text Similarity (STS) is an important task in Natural Language Processing (NLP) applications such as information retrieval, classification, extraction, question answering, and plagiarism detection.
The STS task measures the degree of similarity between two texts and can be expressed as follows: given two sentences, a system returns a continuous score on a scale from 1 to 5, with 1 indicating that the semantics of the sentences are completely independent and 5 meaning that there is a semantic equivalence. 

STS is a difficult issue since languages have numerous ambiguities and synonymous expressions, while
sentences may have variable lengths and complex structures. 
Therefore basic models, e.g. bag-of-words or TF-IDF models, are constrained by their specificities that put aside the role played by the word order and ignore syntactic as well as semantic relationships.
Recent successes in sentence similarity have been obtained using Neural Networks (RNNs: Recurrent Neural Networks \cite{Siamese_LSTM,Kiros,Tai} and CNNs: Convolutional Neural Networks \cite{Similarity_Convolutional}).
Neural Networks (NNs) use a deep analysis of sentences and words to take better into account both the semantics and the structure of sentences in order to predict the sentence similarity.

In this paper, we describe our technique based on NNs to measure similarity. 
First, we use a Siamese CNN to analyze the local context of words in a sentence and to generate a representation of the relevance of a word and its neighborhood.
Then, we use a Siamese LSTM to analyze the entire sentence based on its words and its local contexts.
At last, we predict the semantic similarity of pairs of sentences using the Manhattan distance.

We applied our framework on the SemEval information for STS assignment and we acquired competitive outcomes demonstrating that our model can give helpful information to enhance the sentence analysis.

This paper is organized as follows: we make an overview of relevant work for STS in Section \ref{sc:rw}. 
Next, we detail our approach in Section \ref{sc:oa}. 
The experimental setup and results are presented in Sections \ref{sc:es} and \ref{sc:res}, respectively. 
Finally, we give our conclusion and some last remarks in Section \ref{sc:conc}.

\vspace{-0.5cm}
\section{Related Work}
\label{sc:rw}

To deal with the STS task, previous studies have resorted to various features (e.g. word overlap, synonym/antonym), linguistic resources (e.g. WordNet and pre-trained word embeddings) and a wide assortment of learning algorithms (e.g. Support Vector Regression (SVR), regression functions and NNs).
Among these works, several techniques extract multiple features of sentences and apply regression functions to estimate these similarity scores \cite{lai:2014,zhao:2014,bjerva:2014,Severyn}. 
\citet{lai:2014} analyzed distinctive word relations (e.g. synonyms, antonyms, and hyperonyms) with features based on counts of co-occurences with other words and similarities between captions of images. 
\citet{zhao:2014} predicted the sentence similarity from syntactic relationship, distinctive content similitudes, length and string features. 
\citet{bjerva:2014} also utilized a regression algorithm to foresee the STS from different features (WordNet, word overlap, and so forth). 
Finally, \citet{Severyn} combined relational syntactic structures with SVR.

The development of NNs has improved the results of many NLP applications and especially the STS task \cite{Similarity_Convolutional, Siamese_LSTM, Tsubaki,Rychalska}.
Architectures such as RNNs and CNNs further improve the semantic analysis and the prediction of sentence relatedness.

RNNs differ from other NN models in their ability to process sequential information.
They update a memory cell to make sense of data read in  a sentence over time.
\citet{Rychalska} used a Recursive AutoEncoder (RAE) and a WordNet grant framework to produce sentence embeddings.
They consolidated these embeddings with a Support Vector Machine (SVM) classifier to compute a semantic relatedness score. 
Long Short Term Memory (LSTM) enhances RNNs to handle long-term dependencies \cite{Siamese_LSTM,greff:2015,Tai}.
The LSTM engineering is made out of a memory cell and non-direct gating units that update its state over time and manage the data stream into/out the cell.
\citet{Siamese_LSTM} used a Siamese LSTM to encode sentences using pre-trained word embedding vectors.
Siamese LSTMs used the same weights to encode sentences and to produce comparable sentence representations for similar sentences. 
Then, they predicted the closeness of pair of sentences using the Manhattan distance between the sentence representations. 
\citet{Tai} introduced the Tree-LSTM that is a generalization of LSTM for tree-structured network topologies.
They utilized this Tree-LSTM to encode a couple of sentences and to predict their closeness with a NN that analyzes the distance and the angle between the sentence embeddings.

CNNs have accomplished excellent outcomes in classification \cite{Kim:2014} and other NLP tasks \cite{Collobert:2011}.
\citet{Similarity_Convolutional} generated sentence embedding using a Siamese CNN architecture with various convolution and pooling operations to extract distinctive granularities of information.
Their convolution uses filters that analyze entire word embeddings and each dimension of word embeddings with multiple window sizes.
For output of the convolution operation, they applied several pooling types (max, mean, and min).
Finally, they predicted the sentence similarity from numerous measurements (horizontal and vertical comparison) to compare local regions of sentence representation.

In this work, we join the ideas examined in \cite{Siamese_LSTM} and \cite{Kim:2014} to produce more accurate semantic sentence embeddings. 
The next section presents our model and its characteristics w.r.t. previous work.

\section{Our model}
\label{sc:oa}

A sentence is composed of words which can form phrases and clauses. 
Examining a sentence and its components helps us to comprehend its meaning.
NNs are structures that can inspect relationships between words from multiple points of view. 
On the one hand, LSTMs can recognize and process the semantics of a sentence by investigating the words through time. 
They update their state to get the gist of the sentence (global context) in the order of words.
In this procedure, LSTMs filter unimportant data by retaining just the main information. 
On the other hand, CNNs use layers with convolution filters that are connected to local features \cite{Kim:2014}. 
They enable the analysis of a sentence from multiple perspectives (filters).
This type of NNs does not have the same concern with the sentence length as LSTMs since CNNs examine all the words of the sentence together. 
Nonetheless, CNNs do not consider the order of words in their analysis, so these structures cannot investigate sequence relationships in the sentence.

Differently from \cite{Siamese_LSTM} that only analyze the general context of words and from \cite{Similarity_Convolutional} that do not consider the order of words in the sentences,
we analyze the words in two perspectives: general and local contexts.
Words are considered through time from the general information of a word (word embedding) and its specific semantic and syntactic features (local context) based on its previous and its following words.
We apply a CNN to investigate the local context for each word in a sentence.
The CNN analyzes together all the words of the local context and generates their representation as a unique structure.
Then, we utilize an LSTM to examine the words of the sentence one by one (Figure \ref{img:siamese}).
Our NN has a Siamese structure \cite{Siamese_LSTM,Similarity_Convolutional}, i.e. our $CNN^A$ and our $LSTM^A$ are equal to our $CNN^B$ and our $LSTM^B$, respectively.
The following subsection describes our CNN, our LSTM, and our similarity metrics to predict the sentence similarity.

\begin{figure}[htp]
  \centering
  \includegraphics[width=10cm]{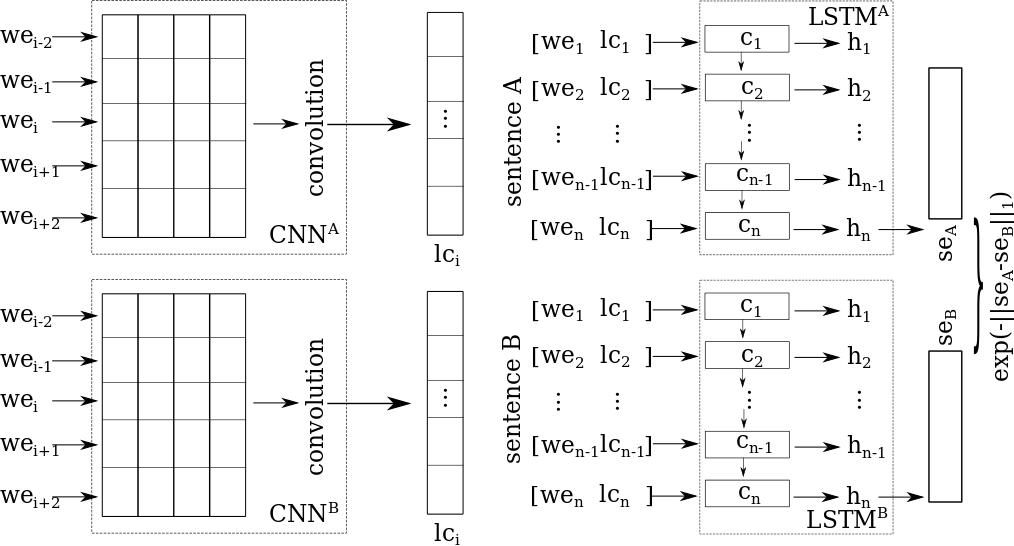}
  \caption {Siamese CNN+LSTM to calculate the similarity of a pair of sentences.}
  \label{img:siamese}
\end{figure}

\vspace{-0.5cm}
\subsection{Neural Network Architecture}
\label{ssc:architecture}

Kim trained a simple CNN on top of pre-trained word vectors for the sentence classification task \cite{Kim:2014}. 
His simple model composed of one layer of convolution achieved excellent results on multiple benchmarks.
Inspired by the good results of CNNs in the sentence classification \cite{Kim:2014}, we use a Siamese CNN to generate local contexts for each word in a sentence from its previous and following words.
We utilize pre-trained word embeddings\footnote{Publicly available at: code.google.com/p/word2vec} to represent these words. 
Let $\mathbf{we}_i \in R^k$ be the $k$-dimensional word vector corresponding to the $i$-th word in a sentence. 
A local context of length $l$ (e.g. $l=5$) is represented as:
\begin{equation}
\mathbf{xl_i} = \mathbf{x_{i-2}} \oplus \mathbf{x_{i-1}} \oplus \mathbf{x_i} \oplus \mathbf{x_{i+1}} \oplus \mathbf{x_{i+2}}
\end{equation}
where $\oplus$ is the concatenation operator.
Our convolution operation involves a filter $w \in R^{lk}$,  which  is  applied  to a window of $l$ words to produce a local context. 
In more details, our CNN generates the local context of word $i$ by:
\begin{equation}
\mathbf{lc_i} = f(\mathbf{w} \cdot \mathbf{xl_i} + \mathbf{b})
\end{equation}

where $\mathbf{b}$ is a bias term and $f$ is the hyperbolic tangent function.
This filter is connected to every sequence of words in a sentence to deliver a local context for all words.

In order to analyze the general and the local contexts of the word $i$, we concatenate its pre-trained word embeddings $\mathbf{we_i}$ (general semantic and syntactic features that were learned on a large corpus) and its local context $\mathbf{lc_i}$.
Our LSTM updates its state $c_i$ and produces an output $h_i$ at time step $i$ in a sentence using the equations described in \cite{Siamese_LSTM}.
The last output of our LSTM $h_n$ represents the meaning of a sentence.

Diverse similarity metrics (cosine, Euclidean and Manhattan distances) were tested and we acquired the best outcome with the Manhattan distance $\textrm{exp}(-||se_A - se_B||) \in [0, 1]$.
Since these scores are not optimized for the similarity metric range (1-5), we apply in a post-processing step a regression method using local regression and bandwidth to project our predictions in the correct scale, similarly to \cite{Li2003}.

\section{Experimental Setup}
\label{sc:es}

We use the SICK dataset to analyze and to test the performance of our system.
This dataset contains 9,927 sentence pairs \cite{sick} and we split it in 4,927/2,000/3,000 for training/validation/test. 
Each sentence pair is annotated with a relatedness label $\in$ [1, 5] corresponding to the average relatedness judged by 10 different individuals. 
The gold scores for relatedness are composed of: 923 pairs within the [1,2) range, 1,373 pairs within the [2,3) range, 3,872 pairs within the [3,4) range, and 3,672 pairs within the [4,5] range.

We initialize our CNN and our LSTM weights with small random Gaussian entries.
Our CNN has filters $R^{300}$ and our LSTM has 50-dimensional hidden representations $\boldsymbol{h}_t$ and memory cells $\boldsymbol{c}_t$. 
We use a forget bias of 2.5 to model long-range dependencies, Adadelta method to optimize the parameters, and a learning rate of 0.01.
We did not identify any improvement with deep LSTMs because of the small amount of data.
Like \cite{Siamese_LSTM}, we also augmented our training dataset and we pre-trained our network using the dataset of SemEval 2013 STS task.

\vspace{-0.3cm}
\section{Results}
\label{sc:res}

In order to understand the relevance of the local context for the sentence similarity, we investigated the original Siamese LSTM without local context and compared it with our method using various lengths for the local context: 3, 5, 7, and 9 (Table \ref{classement}).
The original Siamese LSTM analyzes a sentence considering only the general context of words.
As expected, the analysis of general and local contexts of words improved the sentence analysis, according to the Pearson's and Pearman's correlation coefficients and the Mean Squared Error (MSE) scores.
Short or long local contexts did not generate the best results, which shows that short local context (3 words) did not get enough information about the neighborhood of words and long local context (7 words) includes irrelevant information.

\begin{savenotes}
\begin{table*}[h]
\begin{center}
\begin{tabular}{|l|c|c|c|}
\hline
\textbf{Method} & \textbf{$r$} & \textbf{$\rho$} & \textbf{MSE} \\
\hline
\textit{Siamese LSTM \cite{Siamese_LSTM}}  & \textit{0.8822} & \textit{0.8345} &	\textit{0.2286} \\
Siamese LSTM (publicly available version)\footnote{We used the public version of Siamese LSTM \cite{Siamese_LSTM} available at \url{https://github.com/aditya1503/Siamese-LSTM}, however, we did not get the same results as the ones described in their paper.}  & 0.8500 & 0.7860 & 0.3017 \\
Siamese \#local context: 3 + Siamese LSTM  & 0.8536 & 0.7909 & 0.2915 \\
Siamese \#local context: 5 + Siamese LSTM  & \textbf{0.8549} & \textbf{0.7933} & \textbf{0.2898} \\
Siamese \#local context: 7 + Siamese LSTM  & 0.8540 & 0.7922 & 0.2911 \\
Siamese \#local context: 9 + Siamese LSTM  & 0.8533 & 0.7890 & 0.2923 \\
\hline
Non-Linear Similarity \cite{Tsubaki}                    & 0.8480 & 0.7968 & 0.2904 \\
Constituency Tree LSTM \cite{Tai}					    & 0.8582 & 0.7966 & 0.2734 \\
Skip-thought+COCO (Kiros et al. 2015)                   & 0.8655 & 0.7995 & 0.2561 \\
Dependency Tree LSTM \cite{Tai}							& 0.8676 & \textbf{0.8083} & \textbf{0.2532} \\
ConvNets \cite{Similarity_Convolutional}	            & \textbf{0.8686} & 0.8047 & 0.2606 \\
\hline
\end{tabular}
\end{center}
\caption{\label{classement} Pearson ($r$) and Spearman ($\rho$) correlation coefficients, and Mean Squared Error for the test set of STS task. }
\end{table*}
\end{savenotes}

The bottom part of Table \ref{classement} compares the results of our system and the best state-of-the-art systems.
Although our method did not generate the best results, our system is among the top systems and the results were improved with respect to the publicly available version of the original Siamese LSTM.

In order to illustrate how our local context acts on sentence analysis, Table \ref{tb:local_context} shows at the word level the similarity a pair of paraphrases: \textit{``Her life spanned years of incredible change for women.''} and \textit{``Mary lived through an era of liberating reform for women.''} 
For each pair of words taken in both sentences, the similarity measured as a cosine distance \footnote{The cosine distance between two vectors $u$ and $v$ is defined by $1-\frac{u \cdot v}{||u||_2 ||v||_2}$.} is computed either from general word embeddings (table a) or local contexts of length 5 (table b). The first things to notice is that the two tables have different ranges of values because they each represent a different dimensional space; this means that values must be compared inside each table.
Analyzing Table \ref{tb:local_context}a shows that word embeddings preserve general semantic and syntactic relationships of words.
In this case, the words are more similar to the words that have similar semantics (\textit{1-"Her"}, \textit{2-"Mary"} and \textit{2-"women"}; \textit{1-"life"} and \textit{2-"lived"}; \textit{1-"change"} and \textit{2-"reform"}) and/or have similar syntactic roles (\textit{1-"of"} and \textit{2-"for"}).
Table~\ref{tb:local_context}b highlights that the local context of a word has its semantic and syntactic features based on the words in its window; e.g. the nearest contexts to \textit{1-"life"} are \textit{2-"Mary"}, \textit{2-"lived"}, \textit{2-through} and \textit{2-"women"}  since these local contexts have directly (\textit{2-"lived"}) and indirectly (\textit{2-"Mary"}, \textit{2-"through"} and \textit{2-"women"}) similar semantics.
This analysis is similar to the syntactic features for the local contexts, e.g. the nearest local context of \textit{1-"for"} are \textit{2-"lived"}, \textit{2-"of"}, \textit{2-"for"} and \textit{2-"woman"}.
The relevance of local context is strengthened when we analyze phrasal verbs or multi-word expressions in which meaning depends strongly on their previous and their following words.

\begin{table}[h]
\centering
\begin{tabular}{c}
\resizebox{13.5cm}{!}{
\begin{tabular}{|l|c|c|c|c|c|c|c|c|c|c|}
\hline
& \textbf{Mary}  & \textbf{lived} & \textbf{through} & \textbf{an}    & \textbf{era}   & \textbf{of}    & \textbf{liberating} & \textbf{reform} & \textbf{for}   & \textbf{women} \\ \hline
\textbf{Her}        & 0.77 & 0.93 & 0.90   & 0.81 & 1.04 & 0.92 & 0.95 & 0.91  & 0.80 & 0.80 \\ \hline
\textbf{life}       & 0.91 & 0.70 & 0.89   & 0.90 & 0.82 & 1.00 & 0.71 & 0.86  & 0.88 & 0.86 \\ \hline
\textbf{spanned}    & 0.88 & 0.76 & 0.81   & 1.01 & 0.80 & 0.85 & 0.92 & 1.00  & 0.89 & 0.93  \\ \hline
\textbf{years}      & 0.88 & 0.70 & 0.94   & 0.88 & 0.72 & 0.86 & 0.92 & 0.93  & 0.81 & 0.86 \\ \hline
\textbf{of}         & 0.93 & 0.96 & 0.96   & 1.09 & 0.91 & 0.00 & 0.99 & 1.02  & 0.82 & 0.91 \\ \hline
\textbf{incredible} & 0.94 & 0.89 & 0.83   & 0.94 & 0.84 & 0.95 & 0.74 & 1.04  & 0.83 & 0.97 \\ \hline
\textbf{change}     & 0.97 & 0.90 & 0.93   & 0.92 & 0.85 & 0.99 & 0.80 & 0.67  & 0.83 & 0.92 \\ \hline
\textbf{for}        & 0.96 & 0.97 & 0.67   & 0.79 & 0.89 & 0.82 & 0.88 & 0.92  & 0.00 & 0.89 \\ \hline
\textbf{women}      & 0.81 & 0.96 & 0.99   & 0.93 & 0.92 & 0.91 & 0.79 & 0.88  & 0.89 & 0.00 \\
\hline
\end{tabular}
}
\\
a. Cosine distance between word embeddings. \\
\resizebox{13.4cm}{!}{
\begin{tabular}{|l|c|c|c|c|c|c|c|c|c|c|}
\hline
           & \textbf{Mary}  & \textbf{lived} & \textbf{through} & \textbf{an}    & \textbf{era}   & \textbf{of}    & \textbf{liberating} & \textbf{reform} & \textbf{for}   & \textbf{women} \\ \hline
\textbf{Her}        & 0.06 & 0.08 & 0.09   & 0.11  & 0.16 & 0.12 & 0.13      & 0.13  & 0.09 & 0.08  \\ \hline
\textbf{life}       & 0.10 & 0.08 & 0.09   & 0.12  & 0.11 & 0.13 & 0.13      & 0.14  & 0.10 & 0.10 \\ \hline
\textbf{spanned}    & 0.15 & 0.14 & 0.11   & 0.11  & 0.18 & 0.14 & 0.14      & 0.16  & 0.13 & 0.12 \\ \hline
\textbf{years}      & 0.13 & 0.11 & 0.08   & 0.13  & 0.10 & 0.12 & 0.11      & 0.16  & 0.09 & 0.09 \\ \hline
\textbf{of}         & 0.12 & 0.11 & 0.10   & 0.12  & 0.11 & 0.09 & 0.12      & 0.14  & 0.13 & 0.11  \\ \hline
\textbf{incredible} & 0.12 & 0.12 & 0.13   & 0.14  & 0.19 & 0.13 & 0.03      & 0.16  & 0.14 & 0.09 \\ \hline
\textbf{change}     & 0.14 & 0.13 & 0.18   & 0.15  & 0.18 & 0.15 & 0.16      & 0.02  & 0.15 & 0.13 \\ \hline
\textbf{for}        & 0.10 & 0.09 & 0.10   & 0.11  & 0.12 & 0.08 & 0.11      & 0.12  & 0.04 & 0.08 \\ \hline
\textbf{women}      & 0.09 & 0.07 & 0.09   & 0.11  & 0.11 & 0.08 & 0.09      & 0.14  & 0.07 & 0.01 \\
\hline
\end{tabular}}
\\
b. Cosine distance between local contexts of length 5.
\end{tabular}
\caption{Cosine distance measured between word embeddings (a.) and between the local contexts of length 5 (b.) for each pair of words of two paraphrases.}
\label{tb:local_context}
\end{table}

Table \ref{tb:examples} shows four examples of STS scores for multiple levels of similarities.
The first pair of sentences describes an example of active and passive voice, with the same meaning (4.9 golden score).
The second case is an example of positive and negative sentences (3.3 golden score).
The third example is composed of sentences that do not share the same meaning, having 1.0 golden score.
Finally, our method helps to determine the semantic relationship of the phrasal verb "\textit{wipe off}" and the verb "\textit{clean}" in the last example.
Our approach improves the Siamese LSTM analysis by generating better scores.
The local context helps to better identify not only similar sentences but also the negation and sentences with different meanings.
This local information provides LSTM with a smoother analysis of words and how they connect in a sentence.

\begin{table}[h]
\centering
\begin{tabular}{|p{5.7cm}|c|c|c|}
\hline
\textbf{Pair of sentences} & \textbf{Golden score} & \textbf{Siamese LSTM} & \textbf{Our approach} \\
\hline
\textit{Fish is being cooked by a woman.} & \multirow{2}{*}{4.9} & \multirow{2}{*}{3.84} & \multirow{2}{*}{\textbf{4.05}} \\
\textit{A woman is cooking fish.} &&& \\
\hline
\textit{The bearded man is not sitting on a train.} & \multirow{2}{*}{3.3} & \multirow{2}{*}{3.49} & \multirow{2}{*}{\textbf{3.35}}  \\
\textit{The bearded man is sitting on a train.} &&& \\
\hline
\textit{Someone is playing with a toad.} & \multirow{2}{*}{1.0} & \multirow{2}{*}{1.51} & \multirow{2}{*}{\textbf{1.46}}  \\
\textit{The trumpet is being played by a man.} &&& \\
\hline
\textit{I will wash up if you wipe off the table.} & \multirow{2}{*}{5.0} & \multirow{2}{*}{3.67} & \multirow{2}{*}{\textbf{4.08}}  \\
\textit{I will wash up if you clean the table.} &&& \\
\hline
\end{tabular}
\caption{Examples of semantic textual similarities using Siamese LSTM and our approach (Siamese \#local context: 5 + Siamese LSTM).}
\label{tb:examples}
\vspace{-0.1cm}
\end{table}

To sum up, the local context of words refined the general context analysis.
Our approach identified more details about the words and their local as well as general contexts, which usually leads to improved STS scores.

\section{Conclusion}
\label{sc:conc}
\vspace{-0.1cm}

STS is an important task for various NLP applications, e.g. Automatic Text Summarization (ATS), Question-Answering, Information Retrieval, etc.
Our system combines CNN and LSTM structures to analyze, to identify and to preserve the relevant information in each part of sentences and in the whole sentences.
The local context turned out to be useful to get complement information about a word in a sentence and to improve the sentence analysis.
In our experiments, the local context improved the prediction of the sentence similarity, by reducing the mean squared error and increasing the correlation scores. 

We plan to test other methods to analyze the local context \cite{Ermakova, Zhu}. Unfortunately, the dataset we used for the experiments is of a modest size and we did not find larger annotated corpora for this task. Therefore, we also want to lead extrinsic evaluations by measuring how STS acts on ATS systems, depending on whether the original or the modified Siamese LSTM model is used.

\vspace{-0.1cm}
\section*{Acknowledgments}

This work was partially financed by the European Project CHISTERA-AMIS ANR-15-CHR2-0001.

\bibliographystyle{taln2017}
\renewcommand\bibname{Reference}
\bibliography{biblio}
\end{otherlanguage*}
\end{document}